\documentclass[runningheads]{llncs}
\usepackage[T1]{fontenc}
\usepackage{graphicx}
\usepackage{hyperref}
\usepackage{xcolor}
\hypersetup{
    colorlinks,
    linkcolor={blue!50!black},
    citecolor={blue!50!black},
    urlcolor={blue!50!black}
}
\usepackage{color}
\usepackage{bbm}
\usepackage{amsmath}
\begin{document}
\title{S5CL: Unifying Fully-Supervised, Self-Supervised, and Semi-Supervised Learning Through Hierarchical Contrastive Learning}
\titlerunning{S5CL}
\author{Manuel Tran \inst{1, 2} \and
Sophia J. Wagner \inst{1,2} \and
Melanie Boxberg \inst{1} \and
Tingying Peng \inst{2}}
%\author{Anonymous \inst{1, 2}}
%
\authorrunning{M. Tran et al.}
\institute{Technical University Munich, Munich, Germany \and
Helmholtz AI, Neuherberg, Germany}
%\institute{Institute 1 \and Institute 2}
%
\let\oldmaketitle\maketitle
\renewcommand{\maketitle}{\oldmaketitle\setcounter{footnote}{0}}
\maketitle              % typeset the header of the contribution
\begin{sloppypar} 
\begin{abstract}
In computational pathology, we often face a scarcity of annotations and a large amount of unlabeled data. One method for dealing with this is semi-supervised learning which is commonly split into a self-supervised pretext task and a subsequent model fine-tuning. Here, we compress this two-stage training into one by introducing S5CL, a unified framework for fully-supervised, self-supervised, and semi-supervised learning. With three contrastive losses defined for labeled, unlabeled, and pseudo-labeled images, S5CL can learn feature representations that reflect the hierarchy of distance relationships: similar images and augmentations are embedded the closest, followed by different looking images of the same class, while images from separate classes have the largest distance. Moreover, S5CL allows us to flexibly combine these losses to adapt to different scenarios. Evaluations of our framework on two public histopathological datasets show strong improvements in the case of sparse labels: for a H$\&$E-stained colorectal cancer dataset, the accuracy increases by up to $9\%$ compared to supervised cross-entropy loss; for a highly imbalanced dataset of single white blood cells from leukemia patient blood smears, the F1-score increases by up to $6\%$.\footnote{Code: \url{https://github.com/manuel-tran/s5cl}} 
%\footnote{Code will be made available upon removal of anonymity.} 

\end{abstract}
\section{Introduction}
\label{introduction}
Pixel-wise annotation of histopathological data is highly time-consuming due to the large scale of whole-slide images (WSIs). Therefore, sparse annotations are commonly used~\cite{Bokhorst:2019}, but this results in small amounts of labeled data and an abundance of unlabeled data -- an ideal use case for semi-supervised learning.

Current semi-supervised methods often depend on self-supervised learning which applies domain-specific pretext tasks or contrastive learning techniques to extract useful features from the input data without relying on label information~\cite{Le-Khac:2020,Sahito:2019}. However, the quality of the learned representations highly depends on the artificially designed pretext task \cite{Koohbanani:2021} or the models require large batch sizes, long training times, and high-capacity networks~\cite{Chen:2020b}. Further, the self-supervised models need to be fine-tuned on labeled data in a second step.

Methods that use domain-specific pretext tasks are, for example, COCO~\cite{Yang:2021} and PC-CHiP~\cite{Fu:2020} which apply cross-stain prediction and tissue prediction on histopathological datasets, respectively. SimCLR~\cite{Chen:2020}, BYOL~\cite{Grill:2020}, and Barlow Twins~\cite{Zbontar:2021}, on the other hand, use contrastive techniques to learn feature representations that are invariant to various distortions of the input sample. These methods usually define positive pairs (two differently augmented versions of the same image) and negative pairs (two different images). A contrastive loss then pushes the representations of positive pairs towards each other while pushing the representations of negative pairs away.

Another problem of self-supervised learning is that the extracted features can differ significantly from the ones learned through full-supervision~\cite{Purushwalkam:2020}. Therefore, recent approaches in semi-supervised learning propose to train with both labeled and unlabeled images at the same time. S4L-Exemplar~\cite{Zhai:2019}, for instance, combines a supervised cross-entropy loss on labeled images and a self-supervised triplet loss on the whole dataset. MOAM~\cite{Zhai:2019} and FixMatch~\cite{Sohn:2020} extend this approach by further applying pseudo-labels on unlabeled images. To overcome a confirmation bias towards large classes~\cite{Arazo:2020}, Noisy Student~\cite{Xie:2020} and Meta Pseudo Labels~\cite{Pham:2021} use a teacher-student framework and additionally apply noise or a feedback system, respectively. Yet, these models still have difficulties on highly imbalanced datasets as they are usually biased towards major classes~\cite{Oh:2021}.

\textbf{Contributions.} To solve these issues, we propose a novel framework, called S5CL, that unifies fully-supervised, self-supervised, and semi-supervised learning through hierarchical contrastive learning. We train the model with three contrastive losses simultaneously to construct an embedding space that takes into account the hierarchy of distance relationships between images with respect to their class labels and consistency with different degrees of augmentations. The resulting framework is easy to use and highly flexible: We can omit unlabeled images and train fully-supervised; we can also set the weights of the supervised and semi-supervised loss to zero and train self-supervised; or we train with both labeled and unlabeled images in a semi-supervised way. We validate our method on two public datasets: a colorectal cancer dataset with H$\&$E-stained image patches (NCT-CRC-HE-100K)~\cite{Kather:2019} and a highly imbalanced dataset  of single white blood cells from leukemia patients (Munich AML Morphology)~\cite{Matek:2019}.

\begin{figure}[ht]
\centering
    \includegraphics[width=0.9\textwidth]{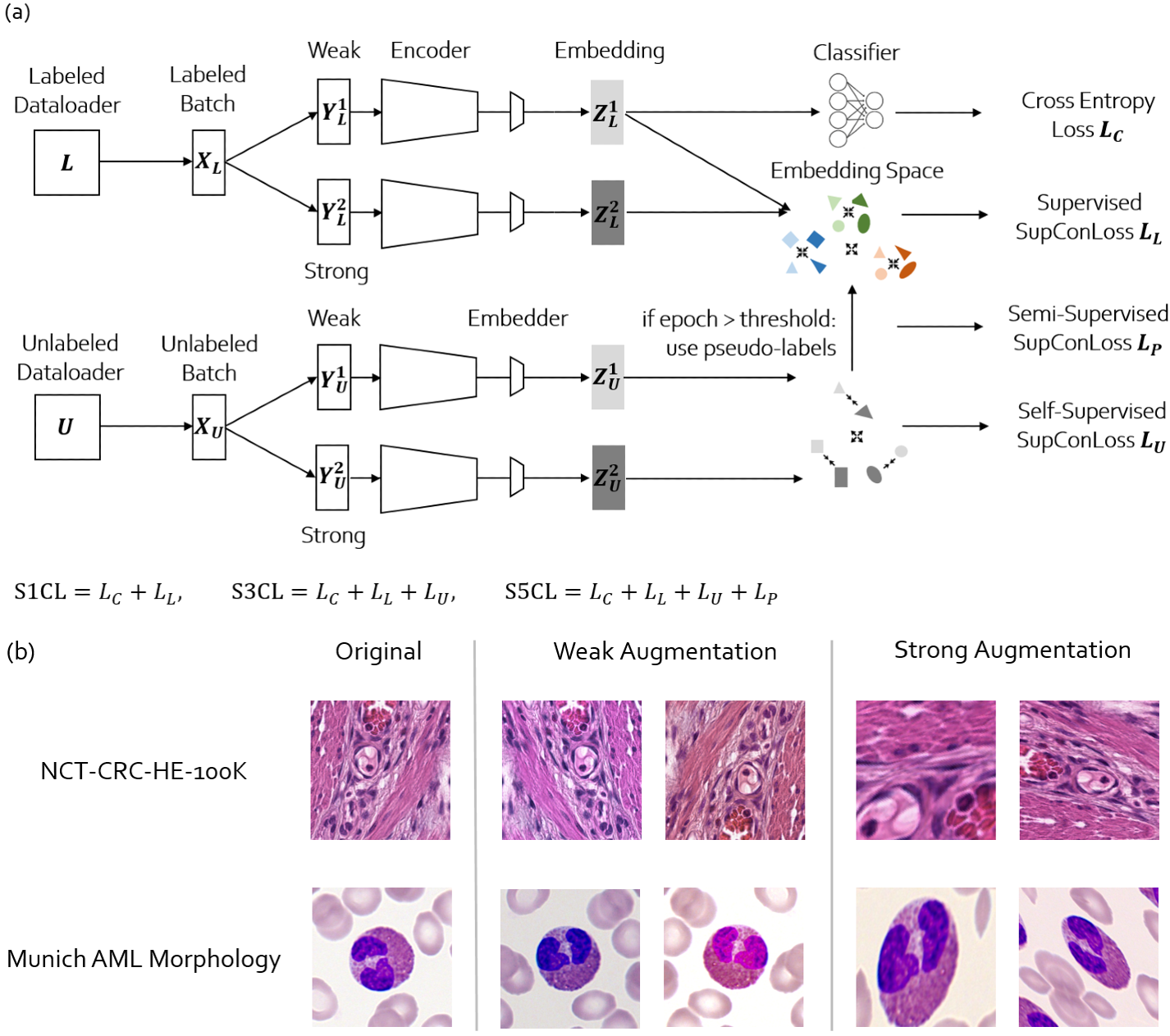}
    \caption{ (a) Overview of S5CL: Colors represent classes; shapes represent instances. Different shades or stretches indicate strongly and weakly augmented data points. (b) Examples of weak and strong augmentations on the two datasets we use.}
    \label{fig:s5cl}
\end{figure}

\section{Method}
\label{method}
S5CL combines three contrastive losses as visualized in Figure \ref{fig:s5cl}a.
First, a supervised contrastive loss embeds images with the same class label into the same cluster and pushes images from other classes away. We improve this by adding weakly and strongly augmented images (see Figure \ref{fig:s5cl}b for some examples) to the loss function (S1CL). Now, weak augmentations and similar images are embedded the closest, followed by different looking images of the same class or strongly augmented views. Images from different classes are the furthest away. Next, a self-supervised contrastive loss additionally pulls the feature representation of unlabeled images and their augmentations towards each other (S3CL). Since both losses act on the same embedding space, the unlabeled image is also forced to have a representation close to its actual but not available class cluster. We further improve the performance by adding pseudo-labels. Instead of applying them to the supervised branch of our method, we add a semi-supervised contrastive loss to circumvent the confirmation bias mentioned above (S5CL).

As contrastive loss, we choose the state-of-the-art SupConLoss \cite{Khosla:2021}. It outperforms other losses like SimCLR~\cite{Chen:2020}, only has one hyperparameter, the temperature, and does not require hard negative mining like the triplet loss. However, our framework is more general and also works with other contrastive losses.

%It only has one hyperparameter, the temperature, that controls the nonlinear mapping from distance metrics to cluster relationships. By using a larger temperature for the self-supervised contrastive loss than for the supervised loss, we can solve the conflicting learning objectives between both losses.

\subsection{Description of S5CL}
\label{description}
Given a labeled dataset $L$, we sample a batch $X_L$ of labeled images. From $X_L$ two batches of distorted images $Y^1_L$ and $Y^2_L$ are created via data augmentation. $Y^1_L$ is weakly augmented with simple color and geometric transformations. It is used for the supervised contrastive loss and the cross-entropy loss.  The second batch $Y^2_L$, on the contrary, is strongly augmented and used for supervised contrastive learning only. Both batches are passed on to the encoder network. The last layer of the encoder is followed by a bottleneck layer called the embedder, producing embeddings $Z^1_L$ and $Z^2_L$. The labeled embeddings are then assessed by the supervised SupConLoss $L_L$~\cite{Khosla:2021}:
\begin{equation}
\label{eq:supconloss}
L_L(\tau)= \sum_{i \in I} \dfrac{-1}{|P(i)|} \sum_{p \in P(i)} \log \dfrac{\exp(\textbf{z}_i \cdot \textbf{z}_p / \tau)}{\sum_{a \in A(i)} \exp(\textbf{z}_i \cdot \textbf{z}_a / \tau)}.
\end{equation}

Here, the anchor $\textbf{z}_i \in Z^j_L$ is an embedding vector with class label $y_i$. As a measure of similarity, SupConLoss calculates the inner product of the anchor with all positive samples in the batch $I$ that belong to the same class, $P(i)=\{p \in A(i) : y_i = y_p\}$. It then applies an exponential function that amplifies large values. The outputs are summed up and normalized over all samples $A=I\setminus\{i\}$. An important hyperparameter is the supervised temperature $\tau=T_L$ that helps distinguishing positive and negative samples. $T_L$ controls the cluster density since intuitively, after training, we have $\textbf{z}_i \cdot \textbf{z}_p > T_L > \textbf{z}_i \cdot \textbf{z}_n $ with $n \in P(i)^C$. 

Similar to the labeled dataset, given an unlabeled dataset $U$, a batch $X_U$ of unlabeled images is sampled and two distorted views $Y^1_U$ and $Y^2_U$ of weakly and strongly augmented images are created. The unlabeled embeddings $Z^1_U$ and $Z^2_U$ are then processed by the self-supervised SupConLoss $L_U$. It has the same form as Equation \ref{eq:supconloss} but with a different temperature $T_U$. In $L_U$, the positive samples $\textbf{z}'_i$ are the different augmented versions of the same image while the negative samples are different images. This could give rise to two contradicting loss objectives: two different samples from the same class are pulled together in the supervised loss $L_L$ (as they belong to the same class) but get pushed away in the self-supervised loss $L_U$ (as they are different samples). In S5CL, we solve this by choosing $T_U>T_L$ and achieve a hierarchy of distance relationships: 
\begin{equation}
\textbf{z}_i \cdot \textbf{z}'_i > T_U> \textbf{z}_i \cdot \textbf{z}_p > T_L > \textbf{z}_i \cdot \textbf{z}_n .
\end{equation}

After a fixed number $t$ of epochs $e$, the classifier can be applied to $Z^1_U$, yielding pseudo-labels. This is passed on to $L_U$. The loss is now semi-supervised and called $L_P$ with temperature $T_P$. The total loss function is thus

\begin{equation}
L_T = \mathbbm{1}_{e<t} \cdot w_U L_U + \mathbbm{1}_{e>t} \cdot w_P L_P + w_L L_L + w_C L_C,
\end{equation}

where $\mathbbm{1}_{e<t}$ and $\mathbbm{1}_{e>t}$ are indicator functions. The positive parameters $w_U$, $w_P$, $w_L$, and $w_C$ are the weights of each loss which are always one in our case. 

%In order to better understand of the contribution of each loss, we also consider two intermediate training targets: i) $w_L L_L + w_C L_C$, which only includes the loss contribution of labeled data, and hence reduces into a fully-supervised method (S1CL); ii) $w_U L_U + w_L L_L + w_C L_C$, which accounts for both self-supervised and supervised contrastive loss whilst neglects the contribution of pseudo-labels (S3CL). 

\subsection{Augmentations}
Each input image is augmented weakly and strongly. Weak augmentations consist of rotations by $0^{\circ}, 90^{\circ}, 180^{\circ}$, and $270^{\circ}$; vertical and horizontal flipping; as well as color jitter (with brightness, contrast, saturation, and hue values all set to 0.1). We always apply rotations and color jitter while flipping is performed with a probability of 0.5. 

Strong augmentations include all augmentations above except that color jitter is replaced by light HED jitter. Additionally, we apply Inception crop for tissue images~\cite{Szegedy:2015}. For single-cell images, we crop between $50\%$ and $100\%$ of the image around the center and re-scale the image to its original size. We lastly employ random affine linear transformations with interpolation (see Figure \ref{fig:s5cl}b).

\subsection{Competitive Methods and Implementation Details}
\label{competitivemethods}
We compare S5CL to the following baseline models: (i) a fully-supervised model that is trained with a cross-entropy loss only (CrossEntropy); (ii) another fully-supervised model that is trained with both a supervised contrastive loss and a cross-entropy loss (SupConLoss); (iii) a state-of-the-art semi-supervised learning method based on a teacher-student network, Meta Pseudo Labels (MPL)~\cite{Pham:2021}, which outperforms other frameworks such as BYOL \cite{Grill:2020} or Noisy Student~\cite{Xie:2020}.

All models use the same encoder -- a ResNet18~\cite{He:2016} pre-trained on ImageNet without the final classification layer. The output of dimension 512 is fed into the embedder which consists of a linear layer without activation and batch normalization layer. The embedder is followed by a linear classifier. MPL uses the same architecture for both student and teacher networks. We employ the following training script for both S5CL and MPL: For each unlabeled batch sampled from the unlabeled dataloader, a labeled batch is sampled from the labeled dataloader.

We tune the hyperparameters, including all temperatures, for each model separately using an internal validation set that differs from the test set. In MPL training, we use Adam with a learning rate of 1e-5 for the student and a learning rate of 3e-5 for the teacher. The weight decay is always 1e-4. On NCT-CRC-HE-100K, the remaining models work well with Adamax while Adam is better on Munich AML Morphology. The learning rate and weight decay are set to 1e-4. %For contrastive learning, we also use a sampler that randomly samples four images per class with replacement for each batch.

\section{Results and Discussions}
\label{results}

\subsection{Evaluation on the Colon Cancer Histology Dataset}
\label{kather}
The NCT-CRC-HE-100K dataset\footnote{\url{https://zenodo.org/record/1214456\#.YhZI1ZYo-Uk}} contains 100,000 non-overlapping image patches from H$\&$E-stained WSIs of colorectal cancer and normal tissue. The images are $224\times224$ pixels large at 0.5 mpp and are color-normalized using Macenko's method. They are annotated into nine well-balanced classes. 

We split the 100,000 images into a training set of 91,000 images and a validation set of 9,000 images (1,000 per class). Next, the original dataset is split into a labeled set containing 5, 10, 20, 50, and 500 images per class which are approximately 0.05$\%$, 0.1$\%$, 0.2$\%$, 0.5$\%$, and 5$\%$ of all labeled images. The remaining images are assigned to the unlabeled set. For the test set, we use CRC-VAL-HE-7K, an independent dataset with 7,180 patches from an external cohort~\cite{Kather:2019}. 

The CrossEntropy and SupConLoss models are trained for 130, 30, and 5 epochs on the sets with 0.05$\%$-0.2$\%$, 0.05$\%$-5$\%$, and 100$\%$ labeled images. Both S5CL and MPL are trained for five epochs. The unlabeled set has a batch size of 128 while the labeled batch size is 32 (for 100$\%$ labeled images, the batch size is also 128). All the embedding spaces have a dimension of 64. The supervised temperature is set to 0.2, and both the self-supervised as well as the semi-supervised temperature are set to 0.7 (see Ablations~\ref{ablations} for details). 

\begin{figure}[t]
\centering
    \includegraphics[width=1.0\textwidth]{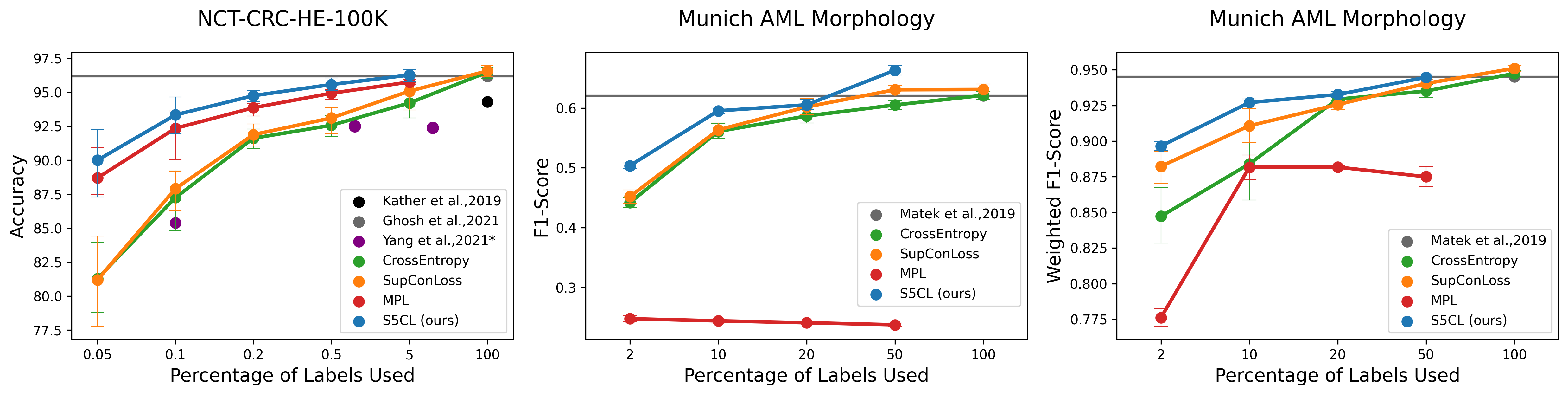}
    \caption{Results for (a) NCT-CRC-HE-100K and (b, c) Munich AML Morphology. Percentages refer to the full dataset for NCT-CRC-HE-100K and to the majority classes in the case of Munich AML Morphology. Models depicted with a gray line are SOTA supervised models in the literature (Ghosh et al.~\cite{Ghosh:2021} and Matek et al.~\cite{Matek:2019}). *Yang et al.~\cite{Yang:2021} is a SOTA semi-supervised baseline on the colon cancer dataset. They train with 8 labels without the background, here we assume 100$\%$ accuracy on that class.}
    \label{fig:results}
\end{figure} 

As shown in Figure \ref{fig:results}, on all reduced labeled sets, S5CL achieves a higher accuracy than the three competitive models including MPL. With only 50 labeled images per class ($0.5\%$), S5CL has an accuracy of $95\%$, which is already comparable to the two reference models in the literature~\cite{Kather:2019} and~\cite{Ghosh:2021} that are trained with the full dataset. With 500 labeled images per class ($5\%$), S5CL almost reaches the accuracy of the full SupConLoss model with $96.6\%$ accuracy. %-- a label efficiency of 20 times. 

\begin{figure}[ht]
\centering
    \includegraphics[width=.9\textwidth]{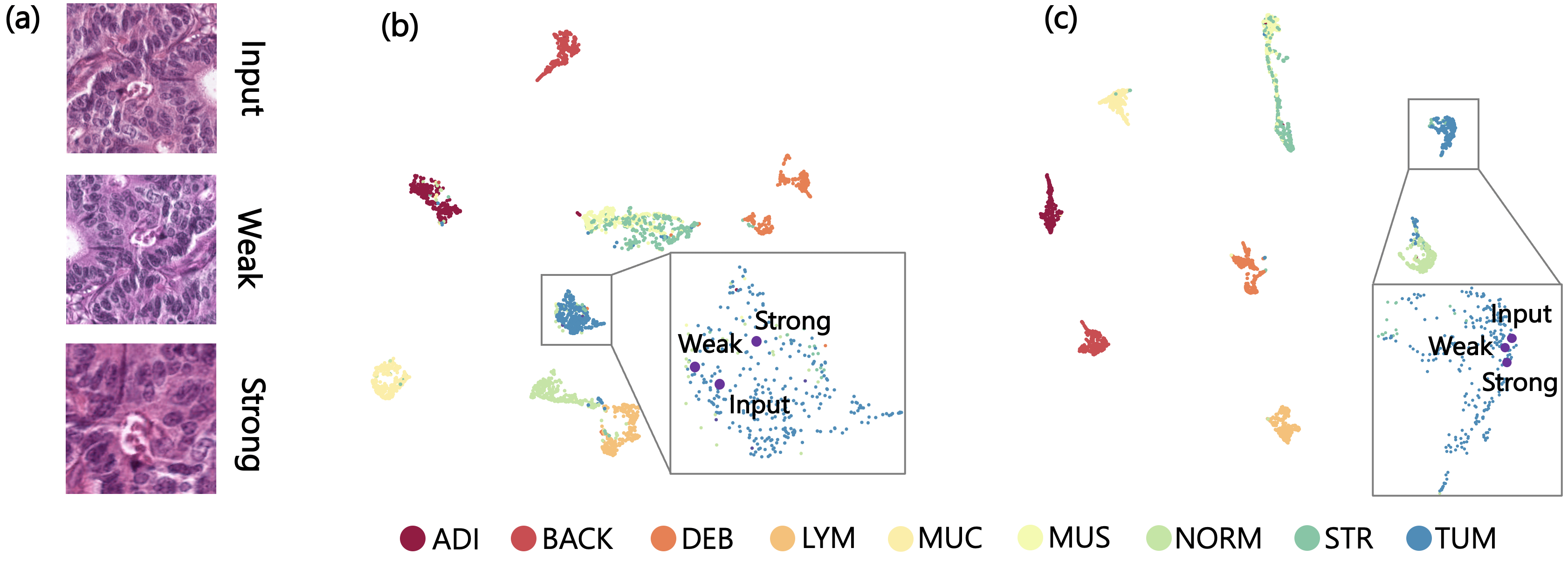}
    \caption{UMAP of feature embeddings for the test set with 300 sampled images per class including weak and strong augmentations using (b) CrossEntropy and (c) S5CL.}
    \label{fig:umap}
\end{figure}

In addition to improving the classification accuracy on small labeled datasets, S5CL also makes the feature embedding space more compact and explicable. To see this, we plot the feature representations as a UMAP in Figure \ref{fig:umap} and measure the embedding quality with the MAP@R score~\cite{Musgrave:2020}, where 100$\%$ implies perfect clustering and separation. S5CL achieves a MAP@R of 74$\%$ whereas CrossEntropy has a score of 67$\%$. In Figure \ref{fig:umap}, we can see that CrossEntropy, might split the representation of the same class (e.g. DEB=debris) into two different clusters --  unlike S5CL. Within a cluster, S5CL also achieves more robust embeddings, i.e., images are ordered hierarchically: weakly augmented images and similar images are embedded the closest to their origins, then comes strong augmentations as well as different looking images from the same class. %For CrossEntropy, even a small permutation like weak augmentation can drive images far away from their origins. 

\subsection{Evaluation on the Leukemia Single-Cell Dataset}
\label{matek}
The Munich AML Morphology dataset\footnote{\url{https://wiki.cancerimagingarchive.net/pages/viewpage.action?pageId=61080958}} contains 18,365 labeled single-cell images taken from peripheral blood smears of 100 patients diagnosed with acute myeloid leukemia (AML). All smears are scanned at $100\times$ magnification with oil immersion at a resolution of 14.14\ mpp and have a patch size of $400\times400$ pixels. The dataset's classes are heavily imbalanced, ranging from 8,484 images in the largest class to 11 images in the smallest class. In addition, some of the 15 cell types are morphologically similar as they are biologically related.

The dataset is split into a training set of size 11,025; a validation set of size 3,666; and a test set of size 3,674. The training set is further split into a labeled and unlabeled set. To preserve the class distribution, we sample 2$\%$, 10$\%$, 20$\%$, 50$\%$, and 100$\%$ images from the large classes for the labeled set and keep all labeled images from small classes that have less than 100 instances per class.

We train the CrossEntropy and SupConLoss models with a labeled batch size of 128; S5CL in addition has an unlabeled batch size of 64. Early stopping is used to avoid overfitting as in~\cite{Matek:2019}. The number of epochs is thus 40, 25, 20, 15, and 15, respectively for each data subset. MPL models use smaller labeled and unlabeled batches of size 32 since training is very memory intensive. However, the batch size does not have a large effect on the results. Due to overfitting, we train the models for two epochs. In addition, we have to include the validation set into the training set, as otherwise, the student model would never see images from the minority class~\cite{Pham:2021}. But we note the class proportions are still the same. For all models, the embedding dimension is 256. The supervised temperature is always set to 0.1 and the self-supervised, as well as the semi-supervised one, are both set to 0.6. 

As shown in Figure \ref{fig:results}, S5CL also outperforms CrossEntropy and SupConLoss for all reduced labeled datasets. Remarkable is that S5CL also exceeds the full models in terms of F1-score, indicating that S5CL is less biased towards the large classes and can also achieve good performance on the minority classes. It is also worth noting that MPL does not perform well on this dataset, which could be due to a confirmation bias towards the majority classes, as described in~\cite{Wei:2021}.

\subsection{Ablation Study}
\label{ablations}
\begin{figure}[t]
\centering
    \includegraphics[width=0.8\textwidth]{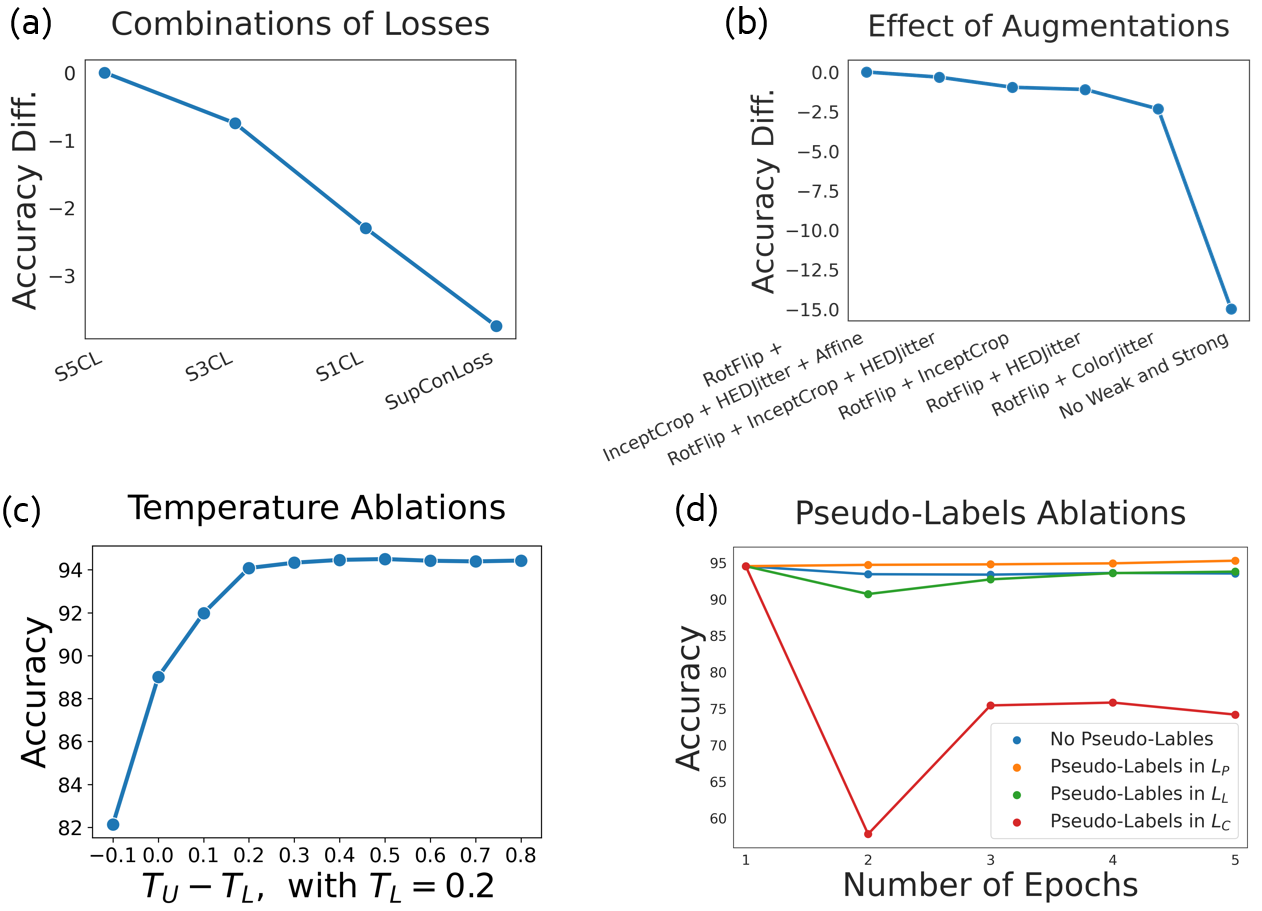}
    \caption{Ablation study for different choices of hyperparameters and architectures.}
    \label{fig:ablations}
\end{figure}

To investigate the contribution of each loss, we first start from a model trained with supervised SupConLoss and progressively build two intermediate models, S1CL and S3CL, to arrive at S5CL (please refer to Figure \ref{fig:s5cl}a). We can see that as the model complexity grows, so does the test accuracy (Figure \ref{fig:ablations}a). Next, we analyze the effect of different augmentation techniques. We start with no augmentations, add weak augmentations and then vary different types of strong augmentations. Our results suggest that both weak and strong augmentations are important and different types of strong augmentations can change the accuracy by $\sim3\%$ (Figure \ref{fig:ablations}b). We also investigate the choice of the self-supervised temperature $T_U$ with respect to the supervised temperature $T_L$. We set $T_L=0.2$ and vary the temperature $T_U$. As expected, a higher distance to $T_L$ resolves the problem of conflicting loss objectives (Figure \ref{fig:ablations}c). The last ablation analyzes the effect of pseudo-labels. If no pseudo-labels are used after the first epoch, the accuracy drops (Figure \ref{fig:ablations}d). Also, if we use pseudo-labels for the classification loss, the accuracy decreases, possibly due to flawed pseudo-labels. 

\section{Conclusion}
We propose S5CL as a unified framework for supervised, self-supervised, and semi-supervised contrastive learning. It combines cross-entropy loss with three contrastive losses for labeled, unlabeled and pseudo-labeled data. In particular, by altering the temperature of the SupConLoss, S5CL can simultaneously train supervised and self-supervised on both labeled and unlabeled data. We evaluate S5CL on two public datasets where it outperforms several baseline methods including Meta Pseudo Labels, a SOTA semi-supervised method. Due to its easy implementation and flexibility, we believe that S5CL can be employed in many problems in computational pathology with sparsely labeled datasets.

%\subsubsection{Acknowledgements} 
%Please place your acknowledgments at the end of the paper, preceded by an unnumbered %run-in heading (i.e. 3rd-level heading).

\bibliographystyle{splncs04}
\bibliography{bibliography}

\end{sloppypar} 
\end{document}